\documentclass[letterpaper, 10 pt, conference]{ieeeconf}  

\IEEEoverridecommandlockouts                              

\usepackage{graphicx} 
\usepackage[utf8]{inputenc}
\usepackage{amsmath,amsfonts,booktabs,cite} 
\usepackage{gensymb} 
\usepackage{siunitx,textcomp} 
\usepackage[hidelinks]{hyperref}
\usepackage{cleveref} 
\usepackage{subcaption}
\usepackage{tabularx}
\usepackage{makecell}
\usepackage{pbox}
\let\labelindent\relax
\usepackage[inline]{enumitem} 
\usepackage[nolist]{acronym} 
\usepackage{multirow}
\usepackage[ruled,linesnumbered, vlined]{algorithm2e}
\usepackage{bm}
\usepackage{titlesec}
\usepackage{balance}
\usepackage[export]{adjustbox}
\usepackage{cellspace}
\usepackage[dvipsnames]{xcolor}

\usepackage[font=small, justification=justified]{caption}

\def\secref#1{Sec.~\ref{#1}}
\def\figref#1{\textbf{Fig.~\ref{#1}}}
\def\tabref#1{\textbf{Tab.~\ref{#1}}}
\def\eqref#1{Eq.~(\ref{#1})}

\newcommand\etal{~\emph{et al. }}

\newsavebox{\twosubbox}

\crefname{algocf}{alg.}{algs.}
\Crefname{algocf}{Algorithm}{Algorithms}

\titleclass{\subsubsubsection}{straight}[\subsection]

\newcounter{subsubsubsection}[subsubsection]
\renewcommand\thesubsubsubsection{\thesubsubsection.\arabic{subsubsubsection}}

\titleformat{\subsubsubsection}
{\normalfont\normalsize\bfseries}{\thesubsubsubsection}{1em}{}
\titlespacing*{\subsubsubsection}
{0pt}{3.25ex plus 1ex minus .2ex}{1.5ex plus .2ex}

\makeatletter
\renewcommand\paragraph{\@startsection{paragraph}{5}{\z@}%
	{3.25ex \@plus1ex \@minus.2ex}%
	{-1em}%
	{\normalfont\normalsize\bfseries}}
\renewcommand\subparagraph{\@startsection{subparagraph}{6}{\parindent}%
	{3.25ex \@plus1ex \@minus .2ex}%
	{-1em}%
	{\normalfont\normalsize\bfseries}}
\def\toclevel@subsubsubsection{4}
\def\toclevel@paragraph{5}
\def\toclevel@paragraph{6}
\def\l@subsubsubsection{\@dottedtocline{4}{7em}{4em}}
\def\l@paragraph{\@dottedtocline{5}{10em}{5em}}
\def\l@subparagraph{\@dottedtocline{6}{14em}{6em}}
\makeatother

\title{\LARGE \bf
NBV-SC: Next Best View Planning based on Shape Completion\\for Fruit Mapping and Reconstruction
}

\author{Rohit Menon \and  Tobias Zaenker \and  Nils Dengler \and Maren Bennewitz
  \thanks{All authors are with the Humanoid Robots Lab, University of Bonn, Germany. Maren Bennewitz is additionally with the Lamarr Institute for Machine Learning and Artificial Intelligence, Germany. This work has partially been funded  by the Deutsche Forschungsgemeinschaft (DFG, German Research Foundation) under Germany’s Excellence Strategy, EXC-2070 -- 390732324 -- Phenorob and under the grant number BE 4420/4-1 (FOR 5351: KI-FOR Automation and Artificial Intelligence for Monitoring and Decision Making in Horticultural Crops, AID4Crops).}
  }

\begin{document}

\maketitle
\thispagestyle{empty} 
\pagestyle{empty}

\begin{abstract} Active perception for fruit mapping and harvesting is a difficult task since occlusions occur frequently and the location as well as size of fruits change over time.
	State-of-the-art viewpoint planning approaches utilize computationally expensive ray casting operations to find good viewpoints aiming at maximizing information gain and covering the fruits in the scene.
	In this paper, we present a novel viewpoint planning approach that explicitly uses information about the predicted fruit shapes to compute targeted viewpoints that observe as yet unobserved parts of the fruits.
	Furthermore, we formulate the concept of viewpoint dissimilarity to reduce the sampling space for more efficient selection of useful, dissimilar viewpoints.
	Our simulation experiments with a UR5e arm equipped with an RGB-D sensor provide a quantitative demonstration of the efficacy of our iterative next best view planning method based on shape completion.
	In comparative experiments with a state-of-the-art viewpoint planner, we demonstrate improvement not only in the estimation of the fruit sizes, but also in their reconstruction, while significantly reducing the planning time.
	Finally, we show the viability of our approach for mapping sweet pepper plants with a real robotic system in a commercial glasshouse.
	
\end{abstract} 

\section{Introduction}
\label{sec:intro}
\begin{figure}[t] 
	\centering
	\includegraphics[width=0.97\columnwidth]{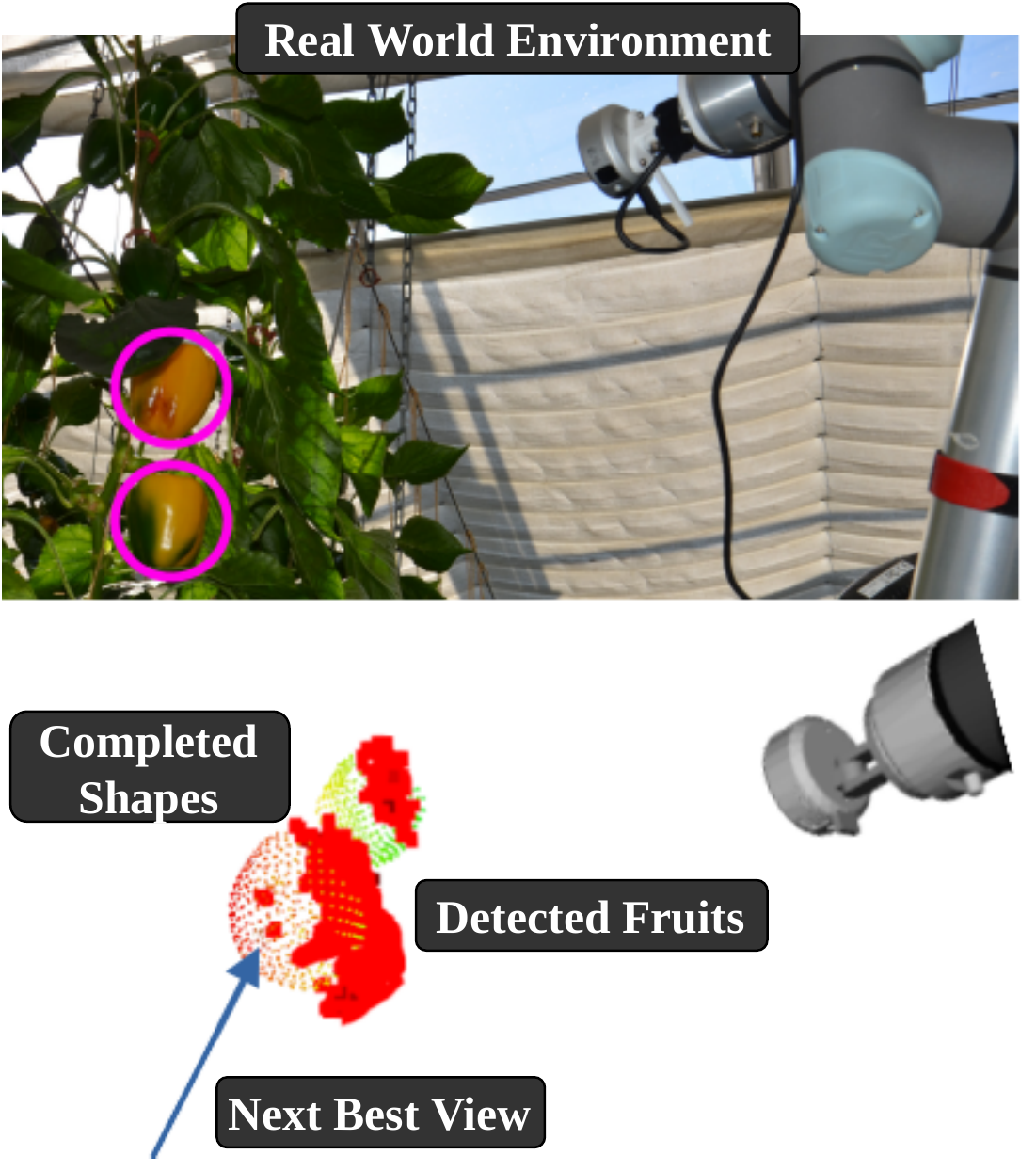} 	
	\captionsetup{width=0.99\columnwidth, justification=justified}
	\caption{Shape completion based viewpoint planning. Here, the red clusters denote the point clouds for the yellow sweet peppers perceived by the Realsense L515 sensor on a UR5e arm. Completed shapes for the partially detected fruits are estimated in real time, which are then used to predict new viewpoints to cover the fruits.} 
	\label{fig:cover}
\end{figure} 

In the context of precision agriculture, fruit detection and mapping is key to yield estimation and harvesting.
Unlike objects found in industrial and household scenarios, fruits and plants change their color, size and shape over time, demanding their spatio-temporal mapping for phenotyping and crop management decisions.
However, frequent occlusion of fruits and the variation in their position on plant structures make their reliable perception a challenging task.
Increasing automation implies the need for autonomous surface inspection methods, which demand complete and accurate fruit mapping, e.g., to plan targeted interventions based on observed quality and ripeness.
State-of-the-art viewpoint planners\cite{zaenker2020viewpoint, burusa2022attention} typically sample random viewpoint candidates and cast rays from the viewpoint origin to the target, to determine the next best view by using occupancy and region of interest~(RoI) information along the ray, which is computationally expensive.
In previous work~\cite{marangoz2022fruit}, we investigated the benefits of shape completion for improving the accuracy of fruit size estimation using data fused from different view poses.
In this paper, we now close the loop by using the shape completion results to inform the viewpoint planner where to plan viewpoints for maximizing the information gain.
\figref{fig:cover} demonstrates that shape completion enables not only to predict the size and position of partially detected fruits, but also to guide the sensor to the next best view using the predicted shapes.
With our approach synthesizing views instead of random sampling, exploiting the missing surface geometry and without any ray casting, we can significantly reduce the planning time.

In more detail, we present NBV-SC, a framework for next best view planning that uses the estimated shape structure of fruits to focus the sensor's attention on the missing surfaces, while favoring viewpoints that are not only aligned with the direction of maximum information gain, but also dissimilar from already performed views.
Quantitative simulation experiments demonstrate the superior performance of our \mbox{NBV-SC~planner} compared to a state-of-the-art viewpoint planner in terms of estimated volume, reconstruction accuracy, and planning time.
Additionally, we carried out qualitative experiments with a real robotic platform measuring sweet pepper plants in a commercial glasshouse to show the applicability of our method under real-world conditions.


To summarize, our contributions are the following: \begin{itemize} 
	\item A novel viewpoint planning approach that uses the missing surfaces of the fruit shapes for finding the next best view.
	\item A formulation of a viewpoint filtering method using viewpoint dissimilarity which improves the computation of the information gain.
	\item A synthesis based approach to viewpoint planning that not only significantly improves the viewpoint planning speed but also leads to better fruit coverage and size estimation.
	
\end{itemize} 

\section{Related Work}
\label{sec:related}
With robots being increasingly deployed outside structured industrial scenarios, active perception is a key factor in improving their efficacy \cite{zeng2020view}.
Viewpoint planning is a subset of active perception where the sensor pose or a sequence of poses is planned to maximize the information gain, i.e., minimize the entropy or uncertainty about the state of the environment or target objects, subject to constraints such as obstacle avoidance and movement cost.
For detail oriented tasks, especially at the object level, such as active recognition, pose estimation \cite{atanasov2014nonmyopic}, mapping or reconstruction \cite{li2005information, kriegel2015efficient}, manipulators or mobile manipulators are typically used with attention-driven next best view (NBV) planning.

In reconstruction tasks, it is typically assumed that the object is not occluded by the environment and, given enough views, can be completely perceived by the viewpoint planning system.
However, fruit mapping is a task where the objects of interest are highly occluded due to fruits growing under the leaves and hence might never be fully reconstructed.
Zaenker\etal \cite{zaenker2020viewpoint} developed RoI targeted viewpoint planning where contours of the detected RoIs are selected as targets for the next best view.
While this approach shows an improvement in volume estimation compared to general viewpoint planners, it only uses the overall contours of the regions of interest without regard for individual shapes, to find the next view, leading to sub-optimal targeting of the missing surfaces of the fruits.
The insertion of regions of interest and ray casting for viewpoint sampling are computationally expensive, leading to performance degradation when handling a large number of fruits.
Burusa\etal \cite{burusa2022attention} demonstrated that placing 3D bounding boxes on different parts of the plant such as stem, leaf nodes, and the whole plant, as an attention mechanism to guide the volumetric NBV planner led to significant improvement in accuracy and speed of reconstruction.
However, the 3D bounding boxes were defined by the user and not autonomously generated,and they also used ray casting for computing the information gain.

In the agricultural context, shape completion has been used for fruit mapping and localization \cite{ge2020symmetry, gong2022robotic, magistri2022contrastive}, without using its output for viewpoint planning.
While Magistri\etal \cite{magistri2022contrastive} provides a contrastive loss based deep learning model for shape completion of pepper fruits, it works on single RGB and depth images and cannot be used with integrated point clouds.
Marangoz\etal \cite{marangoz2022fruit} developed a shape completion approach that used the point cloud obtained from surface mapping\cite{oleynikova2017voxblox} to estimate the position and size of fruits by fitting superellipsoids to the observed shapes, which we extend in our work.
Recently, Schmid\etal\cite{schmid2022scexplorer} demonstrated the application of incremental semantic scene completion for informative path planning of micro-aerial vehicles for efficient environment coverage.
While this work is similar to ours, \mbox{Schmid\etal} focus on efficient coverage of complete scenes and not on the precise mapping of particular objects.

To the best of our knowledge, next best view planning for object mapping and reconstruction based on iterative deformable shape completion has not been carried out till date.
Additionally, unlike our work, most viewpoint planning approaches use computationally expensive ray casting or pixel rendering to compute the information gain.

\section{Our Approach}
\label{sec:approach}
\begin{figure*}[t] \includegraphics[width=\linewidth]{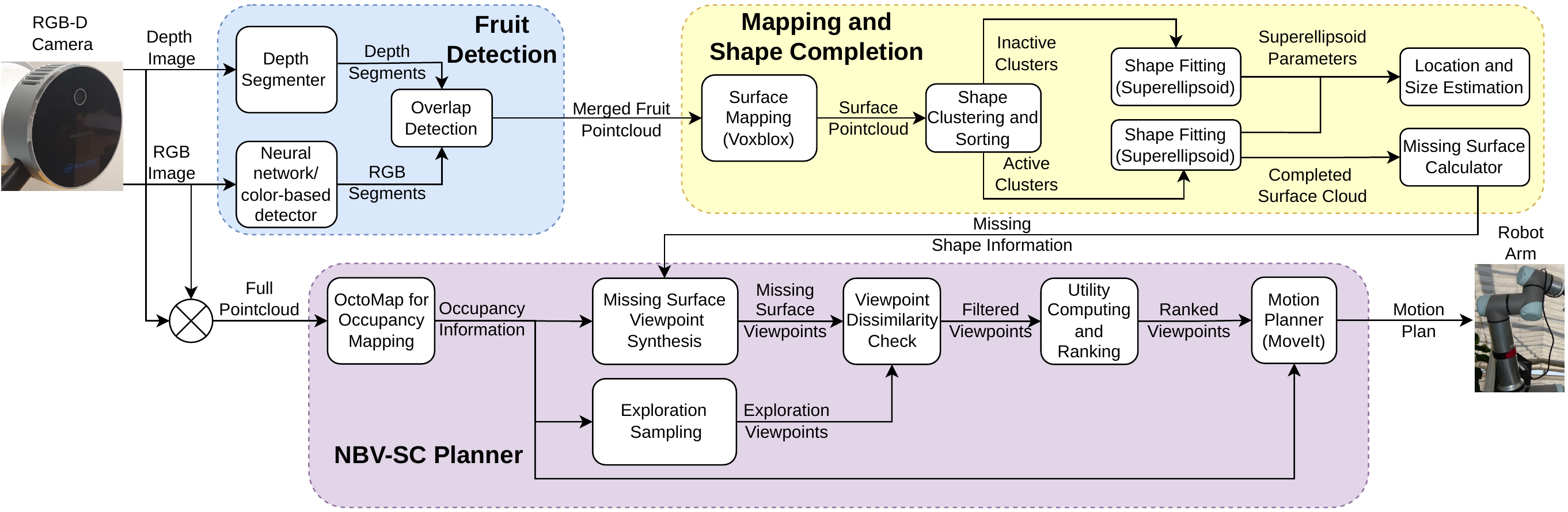} 	\captionsetup{justification=justified} 	\caption{Overview of our system.
		The blue block represents fruit detection (\secref{subsec:roi_detection}), the yellow block represents mapping and shape completion (\secref{subsec:shape_completion}), and the purple block represents the NBV-SC planner (\secref{subsec:viewpoint})} 
	\label{fig:system_overview}
\end{figure*} 

In this paper, we use an RGB-D sensor mounted on a manipulator to plan viewpoints that cover the maximum amount of surfaces of the fruits in the least time while improving the fruit size estimation and reconstruction accuracy.
Our approach consists of three modules: fruit detection, mapping and shape completion, and next best view planning.
At each sensor pose, fruit detection perceives the fruit shapes which are then fed to the mapping and shape completion module. 
The mapping and shape completion module integrates the detected fruit shapes and estimates complete shapes on the mapped fruits.
The estimated shapes are used to plan new views on the missing surfaces, i.e., unobserved parts of the fruits.
\figref{fig:system_overview} gives an overview of our next best view planning approach using shape completion.
The individual steps are described in more detail in the following subsections.

\subsection{Fruit Detection}
\label{subsec:roi_detection}
The color image from the RGB-D camera at each viewpoint is forwarded to an HSV based segmentation method for the red sweet peppers in the simulated scenario. For the real-world experiments, we use a Mask R-CNN \cite{he2017mask} based instance segmentation network, \textcolor{black}{developed by Smitt\etal\cite{mccool21icra},} to detect red, green, and yellow pepper and obtain the fruit masks.
We combine the segments obtained from geometric depth segmentation \cite{furrer2018incremental} with the fruit masks to form the fruits point cloud segments.
These are then combined to form the overall fruit point cloud.

\subsection{Mapping and Shape Completion}
\label{subsec:shape_completion}
The mapping and shape completion module receives the observed fruit point cloud from the fruit detection module at each observation pose, creates an integrated surface map from consecutive observations, and estimates completed shapes for the partially detected fruits.
It runs in parallel to the NBV-SC planner, with its output used for location and size estimation, as well as for planning new viewpoints.

\subsubsection{Surface Mapping} 
\label{subsubsec:mapping}
The merged fruit point cloud at each observation pose is fed to the Voxblox surface mapping system \cite{oleynikova2017voxblox}, which accumulates them over time to form a truncated signed distance fields (TSDF) map.
From the TSDF map, Voxblox outputs the integrated surface point cloud for all detected fruits. In this work, we use a voxel size of 4mm and 64 voxels per side for the surface mapping.
Even though instance-based maps such as Voxblox++\cite{grinvald2019volumetric} are available, we found them to be unreliable as they merged multiple closely positioned fruits into a single instance while at the same time being more computationally expensive.


\subsubsection{Shape Clustering and Sorting} 
We perform Euclidean clustering on the integrated surface point cloud to obtain the different fruit clusters.
We also estimate the centroid of each cluster by computing the surface normals and calculating the least-squares solution to the intersection\cite{marangoz2022fruit}.
We divide the observed shape clusters into two sets: active and inactive.
The active shapes are the ten smallest clusters which we expect to have the most information gain from their predicted shapes, with the remaining ones being the inactive ones.
The sets are continuously updated based on the clustering results.
For the clustering, we use a minimum cluster size of 50, maximum cluster size of 10,000 and cluster tolerance of 1\,cm.

\subsubsection{Shape Fitting} Superellipsoids can be used to describe a variety of shapes and have a closed-form expression for estimating the volume \cite{lehnert2016sweet}.
Therefore, we feed the observed surface point cloud clusters to the superellipsoid matcher \cite{marangoz2022fruit}.
The matcher performs a parametric fitting of the shape by optimizing a cost function that minimizes the deviation in shape center while simultaneously imposing constraints on its dimensions.
The parametric superellipsoid fitting is performed in parallel on the clusters in both the active and inactive sets to estimate fruit centers and size, with the active set executed in a faster loop of 1s and the inactive one in a slower loop of 4s.

\subsubsection{Missing Surface Estimator} 
\label{subsubsec:missing_surface_estimator}
For the active set, we generate a uniformly sampled surface point cloud with a fixed number of points for each completed shape by projecting a Fibonacci sphere on to the fitted superellipsoid and perform a nearest neighbor search for the estimated surface cloud on the observed cloud.
If it is below a minimum (chosen as~1.5\,cm), we remove the point from the estimated surface cloud.
We use a threshold $n_{\mathit{min}}$ for the points remaining in the missing cloud to consider it for viewpoint planning.
The resulting missing surface cloud along with its centroid $c_{\mathit{missing}}$ as well as the optimized center of its superellipsoid shape $c_{\mathit{complete}}$, \textcolor{black}{shown in \figref{fig:missing_viewpoint_synth},} are forwarded to the NBV-SC planner.
\\ 

The advantage of maintaining a separate active set is that both the computation time and memory requirements for maintaining the missing surface information is constant, as the set size remains fixed irrespective of the fruits detected.

\subsection{NBV-SC Planner}
\label{subsec:viewpoint}
As no model of the environment is available, we use a next best view method that performs exploration planning to find new fruits in the unobserved areas, and that subsequently performs targeted missing surface viewpoint synthesis to view the unobserved parts of the already detected fruits.
Furthermore, we formulate a new concept of viewpoint dissimilarity to filter the viewpoints, and compute a missing surface based viewpoint utility. Finally, we rank the viewpoints and execute them using motion planning. In the following sections, the individual steps are described in detail.
\subsubsection{Exploration Sampling}
Initially, our NBV-SC planner performs exploration-based viewpoint planning until it detects fruits and the shape completion module subsequently outputs the missing surface information. 
Also, when there are no longer any useful fruit targeted viewpoints, we switch to exploration sampling.  
To this end, similar to Zaenker\etal\cite{zaenker2020viewpoint} and Monica\etal \cite{monica2018contour}, we perform contour sampling by calculating the frontiers between the occupied and unknown regions using the occupancy information from OctoMap \cite{hornung13auro}. 
At each sensor pose, we update the occupancy information by forwarding the full point cloud without any region of interest information to the OctoMap\cite{hornung13auro} with a default resolution of 1\,cm.

\subsubsection{Missing Surface Viewpoint Synthesis}
\label{subsubsec:synth}
\begin{figure}[t] \includegraphics[width=0.97\columnwidth]{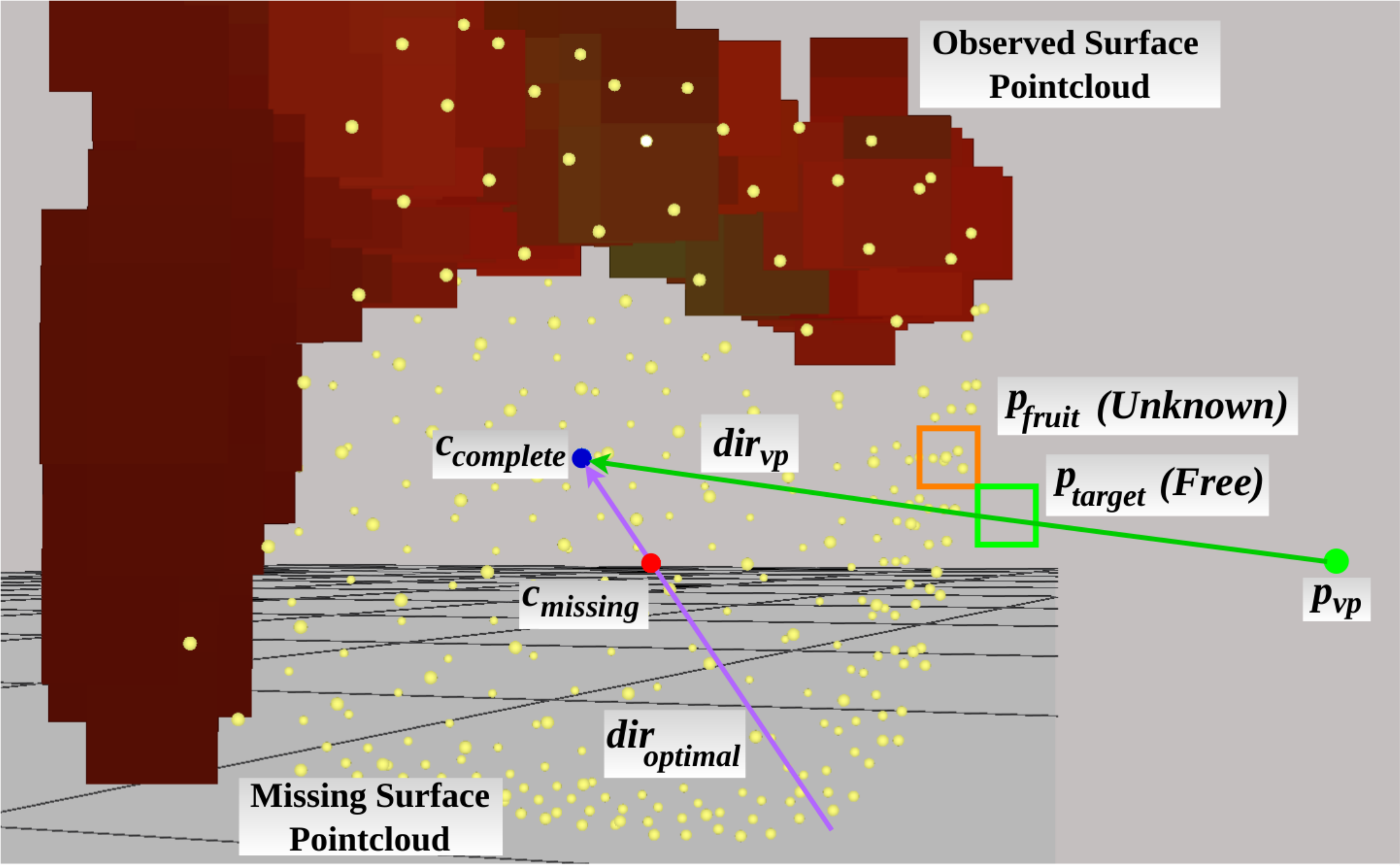} 
	\caption{Missing Surface Viewpoint Synthesis.} 
	\label{fig:missing_viewpoint_synth}
\end{figure}

Unlike the existing viewpoint planners which use random sampling for selecting the viewpoints, we use the geometry of the estimated missing shapes to determine the best directions for targeted viewpoints.
The viewpoint synthesis, with its important variables, is illustrated in \figref{fig:missing_viewpoint_synth}. 
Each point on the missing surface cloud for each shape is checked for its occupancy using OctoMap. It is labeled as a predicted fruit~$p_{\mathit{fruit}}$ node only if the occupancy of the corresponding OctoMap node is unknown.
This is to ensure that we do not label non-fruit occupied structures or free space as $p_{\mathit{fruit}}$ node. 
For every $p_{\mathit{fruit}}$ node, we check its 6 neighborhood (top, bottom, left, right, front, back) for a free node and use it as our viewpoint target~$p_{\mathit{target}}$ to synthesize occlusion-aware targeted viewpoints.
In each iteration, we sample a fixed number of viewpoint targets. 
We then calculate the candidate viewpoint~$vp$ with its direction $dir_{\mathit{vp}}$ given by \eqref{eq:dir_vp} and its origin $p_{\mathit{vp}}$ given by \eqref{eq:origin_vp}.
\begin{equation} \label{eq:dir_vp}
	dir_{\mathit{vp}} = \frac{ c_{\mathit{complete}} - p_{\mathit{target}}}{\lVert  c_{\mathit{complete}} - p_{\mathit{target}}\rVert }
\end{equation}
\begin{equation} \label{eq:origin_vp}
	p_{\mathit{vp}} =  p_{\mathit{target}} - dir_{\mathit{vp}} * rand(d_{min}, d_{max}) 
\end{equation}
where $d_{min}$ and $d_{max}$ are the minimum and maximum distances for sampling the viewpoint origin distance from the target.
The synthesized viewpoint $vp$ is then filtered for reachability, and also for similarity to previous viewpoints, as explained in \secref{subsubsec:vpd}.

\subsubsection{Viewpoint Dissimilarity} 
\label{subsubsec:vpd}
It is better to filter viewpoints that are similar to the already executed ones as they do not usually provide any novel information.
We also observed that the occlusion of fruits by leaves and the limited reachability of manipulators led to the problem of repeated synthesis of similar viewpoints.
To mitigate these problems, we formulated the concept of viewpoint dissimilarity.
In this way, we can eliminate the computation cost of ray casting, by performing this check before computing the information gain.

Each successfully executed viewpoint, whether from missing surface synthesis or exploration sampling, is added to a list of past viewpoints.
During the sampling of new viewpoints, each viewpoint is compared to the past viewpoints.
The origin dissimilarity ${dist}(vp_1, vp_2)$ and angle dissimilarity $\angle(vp_1, vp_2)$ between two viewpoints $vp_1$ and $vp_2$ are formulated as follows: 
\begin{equation}
	\label{eq:similarity_metrics_1}
		{dist}(vp_1, vp_2) = \lVert p_{vp_1} - p_{vp_2}\rVert
\end{equation}
\begin{equation}
	\label{eq:similarity_metrics_1_1}
		\angle(vp_1, vp_2) = arcos(dir_{vp_1} \cdot dir_{vp_2})
\end{equation}
where $p_{vp_1} $ and $p_{vp_2}$ are the respective viewpoint origins, and $dir_{vp_1} $ and $dir_{vp_2}$ the respective viewing directions.

If either $\angle(vp_1, vp_2)$ or ${dist}(vp_1, vp_2)$ is greater than the thresholds $\theta_{\mathit{thresh}}$, $dist_{\mathit{thresh}}$ respectively (chosen empirically as 0.1m and $\pi/6$), the two viewpoints are dissimilar enough and we assign its dissimilarity index $VpD$ as 1.
Otherwise, we calculate the dissimilarity index~$VpD$ as:
\begin{equation}
	\label{eq:vpd_index}
	VpD = (1 - cos(\angle(vp_1, vp_2))) * \frac{{dist}(vp_1, vp_2)}{dist_{\mathit{thresh}}}
\end{equation}
The final $VpD$ is the minimum of the $VpD$ calculated with respect to all the past viewpoints.
If the $VpD$ falls below a threshold (empirically chosen as 0.1), the viewpoint is discarded before computing its information gain and utility, otherwise the information gain is weighted with the dissimilarity index, see below.
\textcolor{black}{$\theta_{\mathit{thresh}}$, $dist_{\mathit{thresh}}$  can be varied to strike a balance between focusing on currently discovered fruits and discovering new fruits. 
Furthermore, lower threshold values lead to more viewpoints being considered for information gain computation leading to more computation time.}

\subsubsection{Viewpoint Utility} For missing surface viewpoint synthesis, we do not perform any ray casting to compute the information gain.
Therefore, we first compute the optimal viewing direction $dir_{\mathit{optimal}}$ for the missing surface, which is the vector from the centroid of the missing surface point cloud $c_{\mathit{missing}}$ to the superellipsoid center $c_{\mathit{complete}}$ as given in \eqref{eq:dir_centroid}: 
 \begin{equation} \label{eq:dir_centroid}
 	dir_{\mathit{optimal}} = \frac{ c_{\mathit{complete}} - c_{\mathit{missing}}}{\lVert  c_{\mathit{complete}} - c_{\mathit{missing}}\rVert }
 \end{equation}
We then use the data from the missing surfaces to estimate the information gain $IG_{\mathit{missing}}$ of a viewpoint $v_p$ as follows: 
\begin{equation}
 IG_{\mathit{missing}} = \frac{n_{\mathit{missing}}}{n_{\mathit{min}}}*(dir_{\mathit{vp}} \cdot dir_{\mathit{optimal}}) *  VpD	
\end{equation}
where $n_{\mathit{missing}}$ is the number of points in the missing surface cloud, which is proportional to its surface area, $n_{\mathit{min}}$ is the minimum number of surface points needed to consider it for the planner as mentioned in \secref{subsubsec:missing_surface_estimator}.
The term $n_{\mathit{missing}} \backslash n_{\mathit{min}}$ represents the thresholded information gain of the considered shape, whereas $dir_{\mathit{vp}}*dir_{\mathit{optimal}}$ represents the alignment of the current viewpoint to the optimal viewpoint direction.

The overall utility of the viewpoint, with the camera currently at position $p_{\mathit{cam}}$, is calculated as follows: 
\begin{equation}
	\label{eq:utility}
	U_{\mathit{missing}} = IG_{\mathit{missing}} - \alpha * \lVert p_{\mathit{vp}} - p_{\mathit{cam}} \rVert
\end{equation}
where $\lVert p_{\mathit{vp}} - p_{\mathit{cam}} \rVert$ represents the motion cost in Euclidean space and $\alpha$ is a scaling factor set to 0.2 in our experiments.
For exploration sampling, the standard entropy-based information gain $IG_{\mathit{exp}}$ is computed, using ray casting:
\begin{equation}
	\label{eq:IG_EXP}
	IG_{\mathit{exp}} = \frac{1}{n_{rays}}\sum{\frac{n_{\mathit{unknown}}}{n_{\mathit{all}}}}
\end{equation}
where $n_{\mathit{unknown}}$ are the unknown voxels encountered by the ray cast from the viewpoint origin to the target, $n_{\mathit{all}}$ are the total voxels encountered by the ray and $n_{rays}$ are the number of rays cast.
The utility $U_{\mathit{exp}}$ is calculated taking into account the motion cost, similar to \eqref{eq:utility}.

\subsubsection{Viewpoint Selection and Execution} If we have missing surface information available, we first perform targeted viewpoint synthesis as described in \secref{subsubsec:synth}.
Only those viewpoints that have a utility score above a certain threshold (set as 0.2) are added to the list of selected viewpoints, which are sorted according to their utility value. 
If the list is empty, we switch to exploration sampling.
\textcolor{black}{The threshold for the utility of targeted viewpoints can be varied to balance between exploratory and targeted viewpoints.}
The viewpoints are serially forwarded to the MoveIt planner \cite{moveit} for collision-free motion planning and trajectory execution, until a successful execution is performed.

\section{Experimental Evaluation}
\label{sec:exp}
The goal of our experimental evaluation is to provide a quantitative comparison of the performance of our NBV-SC planner with a state-of-the-art planner in different scenarios.
The evaluation is carried out with respect to the mapping performance, reconstruction accuracy, and viewpoint planning time.

\subsection{Experimental Setup}
\begin{figure}[t]
	\captionsetup{justification=justified}	
	\begin{subfigure}[t]{0.32\columnwidth} 		\centering 		\includegraphics[width=\columnwidth]{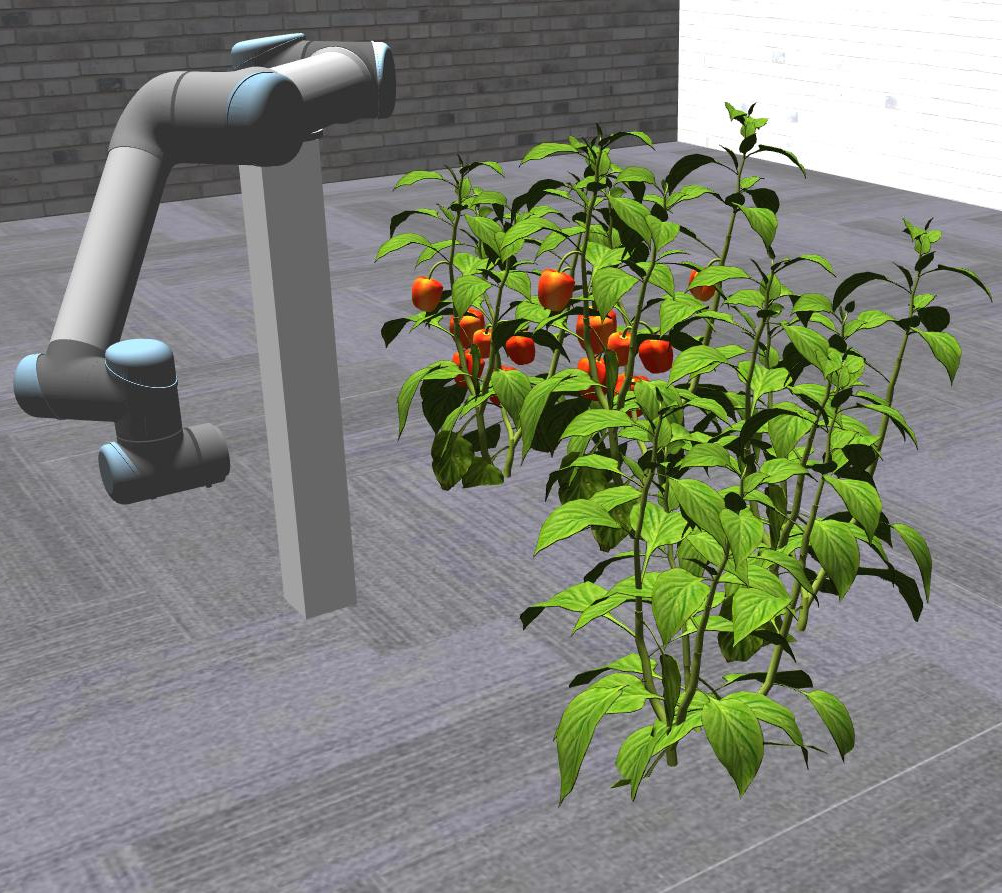} 		\caption{Scenario 1} 
		\label{fig:simulatedenv1}
	\end{subfigure} 	
	\begin{subfigure}[t]{0.32\columnwidth} 		\centering 		\includegraphics[width=\columnwidth]{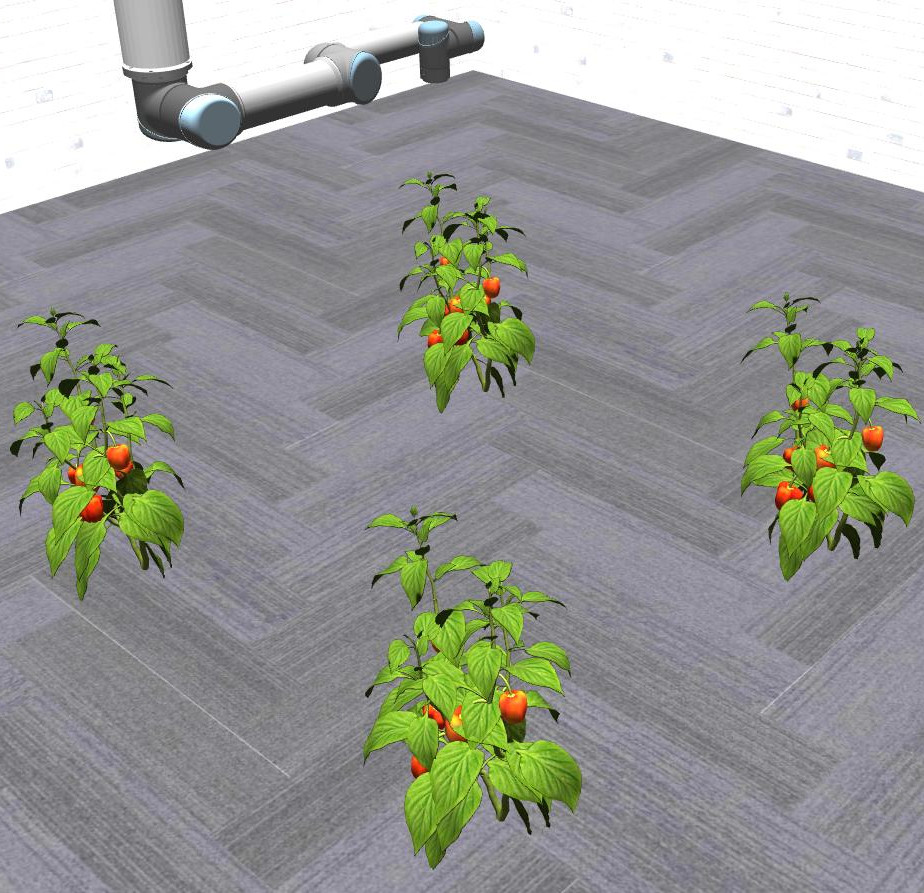} 		\caption{Scenario 2} 
		\label{fig:simulatedenv2}
	\end{subfigure} 	
	\begin{subfigure}[t]{0.32\columnwidth} 		\centering 		\includegraphics[width=\columnwidth]{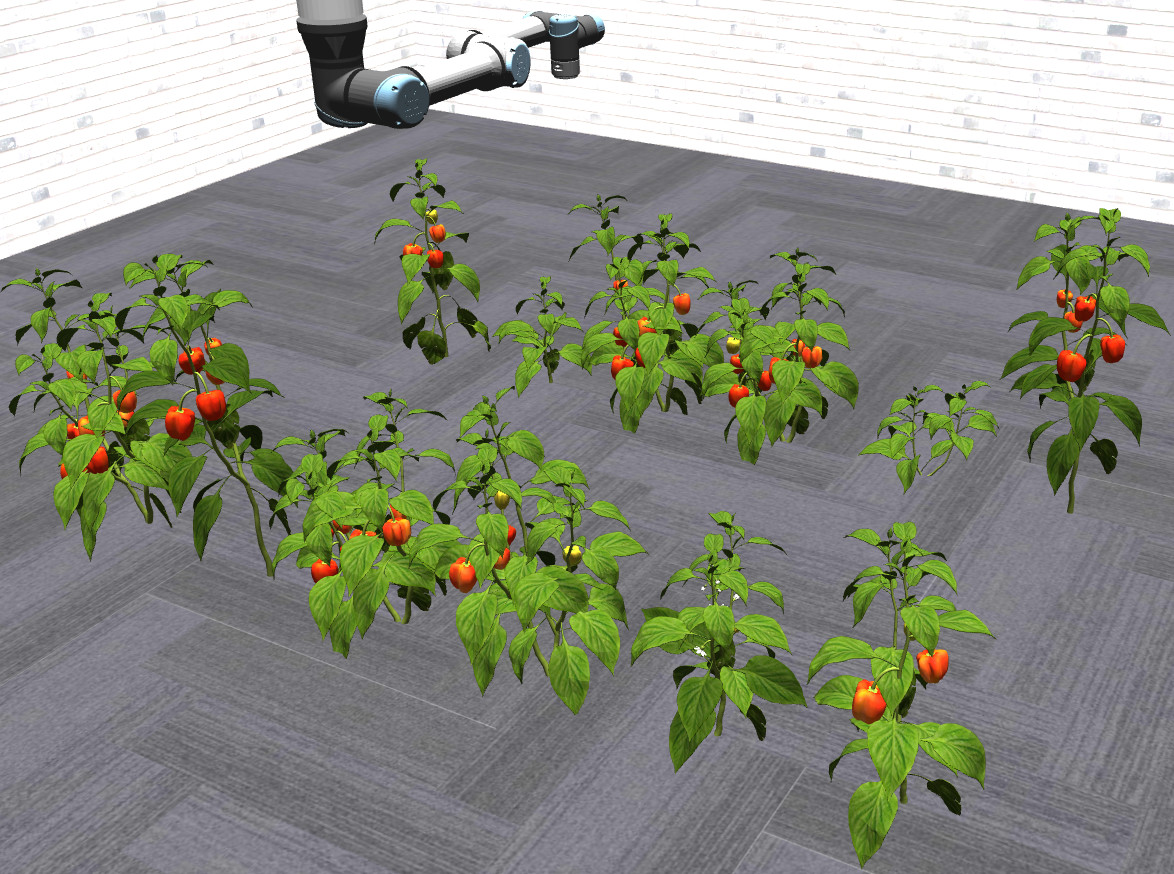} 		\caption{Scenario 3} 
		\label{fig:simulatedenv3}
	\end{subfigure} 
		\caption{Simulation scenarios used for the experimental evaluation. \textit{Scenario 1:}  Environment with a static arm, as used in~\cite{zaenker2020viewpoint}, with 4 plants and 14 fruits.
		\textit{Scenario 2}: Environment with a retractable arm as in \cite{zaenker2020viewpoint} with 4 plants and 28 fruits.
		\textit{Scenario 3}: Environment with a retractable arm as in \cite{marangoz2022fruit} with 12 plants and 54 fruits. 
	    \textcolor{black}{The fruits are 7-9 cm long and wide.}}
	\label{fig:simulation_env}
\end{figure}
We carried out our simulation experiments using the Gazebo framework\cite{koenig2004design} on a computer with core i7-1200H processor, 32GB RAM and RTX3060 GPU with ROS Noetic.
We compare the performance of our \textbf{NBV-SC} planner to the RoI viewpoint planner (\textbf{RVP})~\cite{marangoz2022fruit}, which uses viewpoint sampling based on regions of interest~\cite{zaenker2020viewpoint}, and shape completion only for position and volume estimation.
Additionally, we added the concept of viewpoint dissimilarity to the RVP, to understand its effect on the existing planner, which we denote as \textbf{RVP+VPD}.
For the evaluation, we used the three scenarios shown in \figref{fig:simulation_env}.
We carried out 10 trials in each scenario and use a plan length of 120\,s for each trial, where the plan length corresponds to the trajectory execution duration.
We simulated depth noise for the RGB-D camera by adding Gaussian noise with a standard deviation of 0.003 and salt and pepper noise for zero pixels with a probability of 0.003.
We also added shadowing effect to the depth image.

\subsection{Fruit Detection and Size Estimation}

We compare the performance of our NBV-SC planner to RVP and RVP+VPD with respect to the detection of fruit clusters and the volume accuracy of their fitted shapes.
To this end, we use the ground truth fruit shapes from the simulation and perform superellipsoid fitting on them to compute their centers and volumes.
A superellipsoid fitted on an observed shape is considered to be successfully detected, if the distance between its center and its corresponding ground-truth shape center is less than 10\,cm.
The volume accuracy of a completed fruit shape is then computed as follows: 
\begin{equation}
	\label{eq:vol_acc_eqn}
	acc_{V} = 1 - \frac{|V_{det} - V_{gt}|}{V_{gt}}
\end{equation}
where $V_{det}$ is the detected volume and $V_{gt}$, the ground truth volume.

\begin{figure*}[t] 	\begin{subfigure}[t]{0.32\linewidth} 	\raggedleft	 		\includegraphics[width=\linewidth,right]{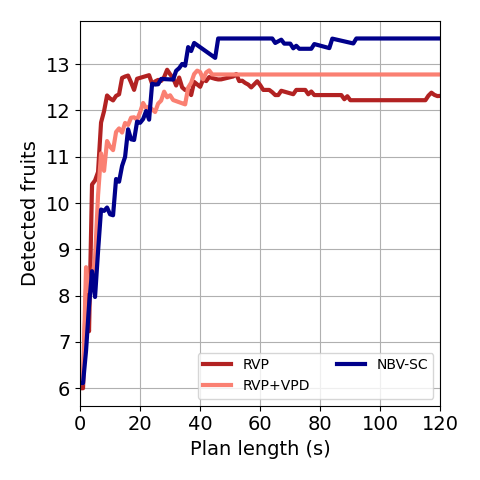} 		\includegraphics[width=\linewidth,right]{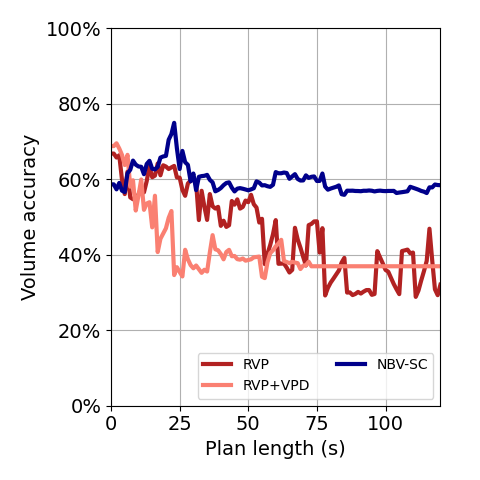}	\caption{Scenario 1} 
		\label{fig:s1_clusters}
	\end{subfigure} 	\begin{subfigure}[t]{0.32\linewidth} 		\raggedleft	 	 	 		\includegraphics[width=\linewidth,right]{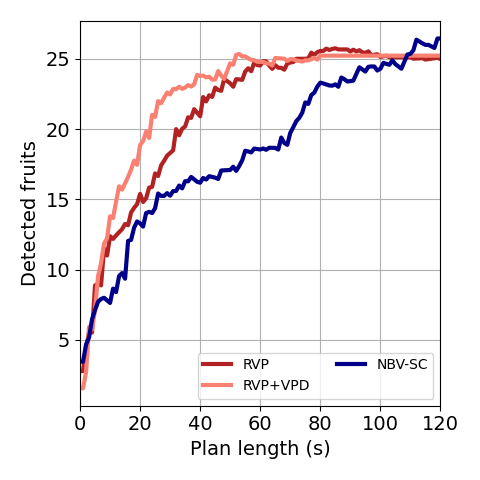} 	\includegraphics[width=\linewidth,right]{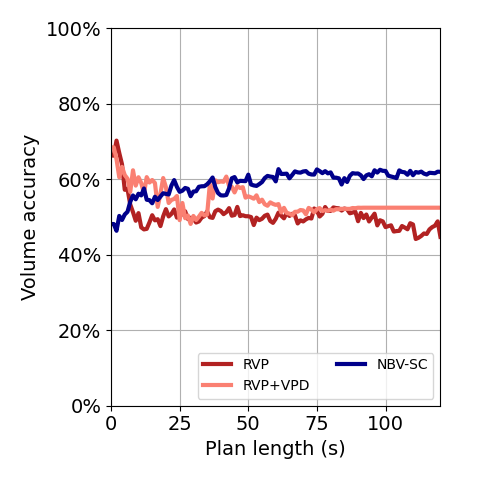}	\caption{Scenario 2} 
		\label{fig:s2_clusters}
	\end{subfigure} 	\begin{subfigure}[t]{0.32\linewidth} 	\raggedleft	 	 		 		\includegraphics[width=\linewidth,right]{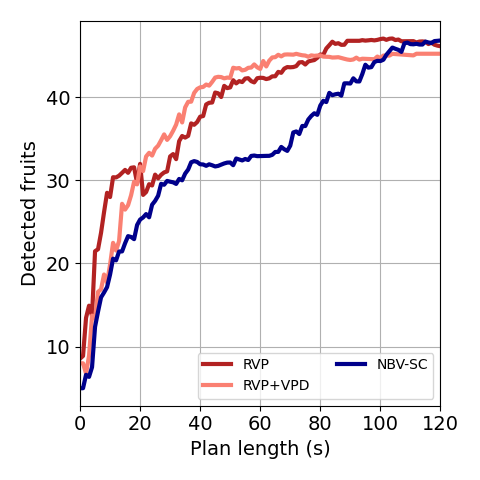} \includegraphics[width=\linewidth,right]{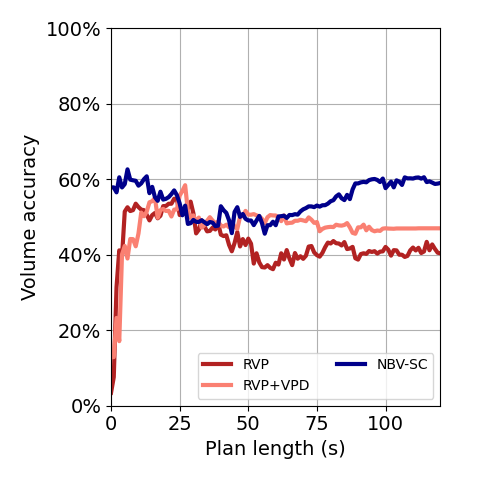} 		\caption{Scenario 3} 
		\label{fig:s3_clusters}
	\end{subfigure} 	\caption{Number of detected fruits and their mean volume accuracy over plan length in seconds.
		NBV-SC denotes our shape completion based next best view planner, RVP denotes the RoI Viewpoint planner with shape estimation~\cite{marangoz2022fruit}, and RVP+VPD denotes RVP with the viewpoint dissimilarity check added.
		NBV-SC is slightly better than RVP and RVP+VPD in number of detected fruits~(top), however, \mbox{NBV-SC}  outperforms RVP and RVP+VPD considerably in the accuracy of volume estimation of completed shapes~(bottom).
		NBV-SC plans more views on individual fruits leading to better individual coverage.
		Volume accuracy is the accuracy of the completed shape's volume compared to the ground-truth shape's volume calculated as in \eqref{eq:vol_acc_eqn}.} 
	\label{fig:results_num_vol}
\end{figure*} 

\figref{fig:results_num_vol} shows the results for fruit detection and volume accuracy for all planners in the three scenarios.
As can be seen, in all the scenarios, our NBV-SC planner achieves a  similar or higher number of correctly detected fruits at the end of the evaluation compared to the baselines.
Our \mbox{NBV-SC} planner detects new fruits only after finishing viewing the current set of fruits for which missing surface information is high.
This phenomenon can be seen in Scenarios 2 and 3, where the number of fruits detected by NBV-SC remains almost constant in the middle.
RVP and RVP+VPD, on the other hand, do not perform enough targeted viewing of fruits since they do not use the missing surface information during view pose sampling.
Hence, they switch to exploration sampling frequently and detect new clusters faster initially.
This, however, comes at the cost of reduction in volume estimation accuracy.

\begin{table}[t] \centering 
	\resizebox{\columnwidth}{!}{
		\renewcommand{\arraystretch}{1.1}%
		\begin{tabular}{l|r|r|r}
			Planner& Scenario 1     			& Scenario 2               & Scenario 3       \\ \hline
			NBV-SC   &$\mathbf{58.5} \pm 4.42$  & $\mathbf{62.0} \pm 1.58$ & $\mathbf{59.1} \pm 3.16$ \\
			RVP      &$32.0 \pm 17.4$ 		    & $44.8 \pm 3.79$          & $40.3 \pm 4.74$ \\
			RVP+VPD  &$36.9 \pm 12.0$ 			& $52.5 \pm 4.10$           & $47.0 \pm 3.79$
		\end{tabular}
	}
	\caption{Mean volume estimation accuracy with standard error, at the end of the evaluation.  
		The NBV-SC planner achieved significantly better results ($p<0.05$) compared to RVP~\cite{marangoz2022fruit} in all three scenarios.} \label{tab:vol_accuracy}
\end{table}

\textcolor{black}{In all the three scenarios, the actual distance of the superellipsoid centers to the ground truth centers is less than 2.5 cm for all the three planners.} The results of the volume estimation accuracy are shown in the lower part of \figref{fig:results_num_vol}, and numerically in \tabref{tab:vol_accuracy}.
Our NBV-SC planner outperforms both baselines with respect to volume estimation accuracy in all the three scenarios.
NBV-SC performs more targeted observations of the already detected fruits which enables it to improve the volume estimation.
The main reason for the lower accuracy of RVP is that fruits get re-detected from larger distances during exploration sampling, which leads to depth perception errors and merging of fruit clusters.
Pure RVP's volume accuracy is also slightly lower than that of RVP+VPD as it stops exploration sampling when no new dissimilar viewpoints are found.

\subsection{Reconstruction Accuracy}
\begin{table}[t] \centering 
	\resizebox{\columnwidth}{!}{
	\renewcommand{\arraystretch}{1.1}%
	\begin{tabular}{l|r|r|r}
		Planner& Scenario 1     & Scenario 2      & Scenario 3       \\ \hline
		NBV-SC   & $\mathbf{6.9} \pm 0.16$  & $\mathbf{7.0} \pm 0.03$ & $\mathbf{6.8} \pm 0.10$ \\
		RVP      & $8.7 \pm 0.32$  & $7.7 \pm 0.13$ & $7.7 \pm 0.13$  \\
		RVP+VPD  & $8.0 \pm 0.35$  & $7.9 \pm 0.06$ & $7.8 \pm 0.16$  
	\end{tabular}
}
\caption{ Mean Chamfer distances (in mm) with the standard error, of the detected point clouds generated from our voxblox surface map, obtained after a plan length of 120\,s or the time at termination of sampling, compared to the ground truth.
	NBV-SC  achieved significantly better results ($p<0.05$) compared to RVP~\cite{marangoz2022fruit} in all the three scenarios and compared to RVP+VPD in Scenarios 2 and 3.} \label{tab:chamferdist}
\end{table}
We additionally calculated the Chamfer distance\cite{borgefors1983chamfering} of the generated fruit point cloud from the surface map to the ground truth to evaluate the final quality of our reconstructed fruit meshes.
\tabref{tab:chamferdist} shows that with our approach, the average distances are 1-1.5\,mm more accurate compared to baselines.
While this is only a small improvement, it applies consistently across all scenarios.
Our planner outperforms RVP in all scenarios and RVP+VPD in Scenarios 2 and 3 statistically significantly, according to the one-sided MannWhitney U test with $p<0.05$.
Our method generates more viewpoints per fruit, which leads to this improvement.
It is likely that in scenarios with more occluded regions and more sensor noise, this behavior could prove to be even more beneficial.
\subsection{Viewpoint Planning Time}
To demonstrate the strength of our approach, we finally performed a run time analysis.
Viewpoint planners, such as RVP, that use ray casting for calculating the information gain are computationally expensive and consume a significant amount of time for viewpoint sampling.
As can be seen in \tabref{tab:vpt}, RVP requires more than 2\,s per iteration of sampling of 100 viewpoints in Scenarios 2 and 3.
In contrast, NBV-SC reduces this time by 80-90\% in the same scenarios.
We achieve this significant reduction in planning time by synthesizing viewpoints using the missing surface information, rather than search-based sampling for targeted viewpoints.
Due to the low number of targets in Scenario~1, the time reduction is not as significant, since NBV-SC performs more frequent exploration sampling.

It can also be seen that even though RVP+VPD is significantly worse than NBV-SC in the second and third scenarios, it still performs better than RVP, confirming our hypothesis that viewpoint dissimilarity can reduce the planning time for ray casting based viewpoint planners as well.
\begin{table}[t] \centering 
	\resizebox{1.01\columnwidth}{!}{
	\renewcommand{\arraystretch}{1.2}%
	\begin{tabular}{l|r|r|r}
		Planner & Scenario 1 				& Scenario 2 & Scenario 3 \\ \hline
		NBV-SC  &  $\mathbf{0.70} \pm 0.23$ &  $\mathbf{0.27} \pm  0.20$           & $\mathbf{0.50} \pm0.27$       \\
		RVP     &  $0.78\pm 0.35$           & $2.45 \pm 0.44$           &$2.52 \pm 0.49$            \\
		RVP+VPD &  $0.89 \pm  0.36$         &  $1.37 \pm  0.28$           & $1.87 \pm 0.42$           
	\end{tabular}
}
	\caption{Mean viewpoint planning time (seconds) with standard error per iteration.
		NBV-SC achieved significantly better results~($p<0.05$) compared to RVP~\cite{marangoz2022fruit} and RVP+VPD in Scenarios~2 and 3.} \label{tab:vpt}
\end{table}

\subsection{Real-World Experiments}
\begin{figure}[t] 
	\centering
	\includegraphics[width=0.99\columnwidth]{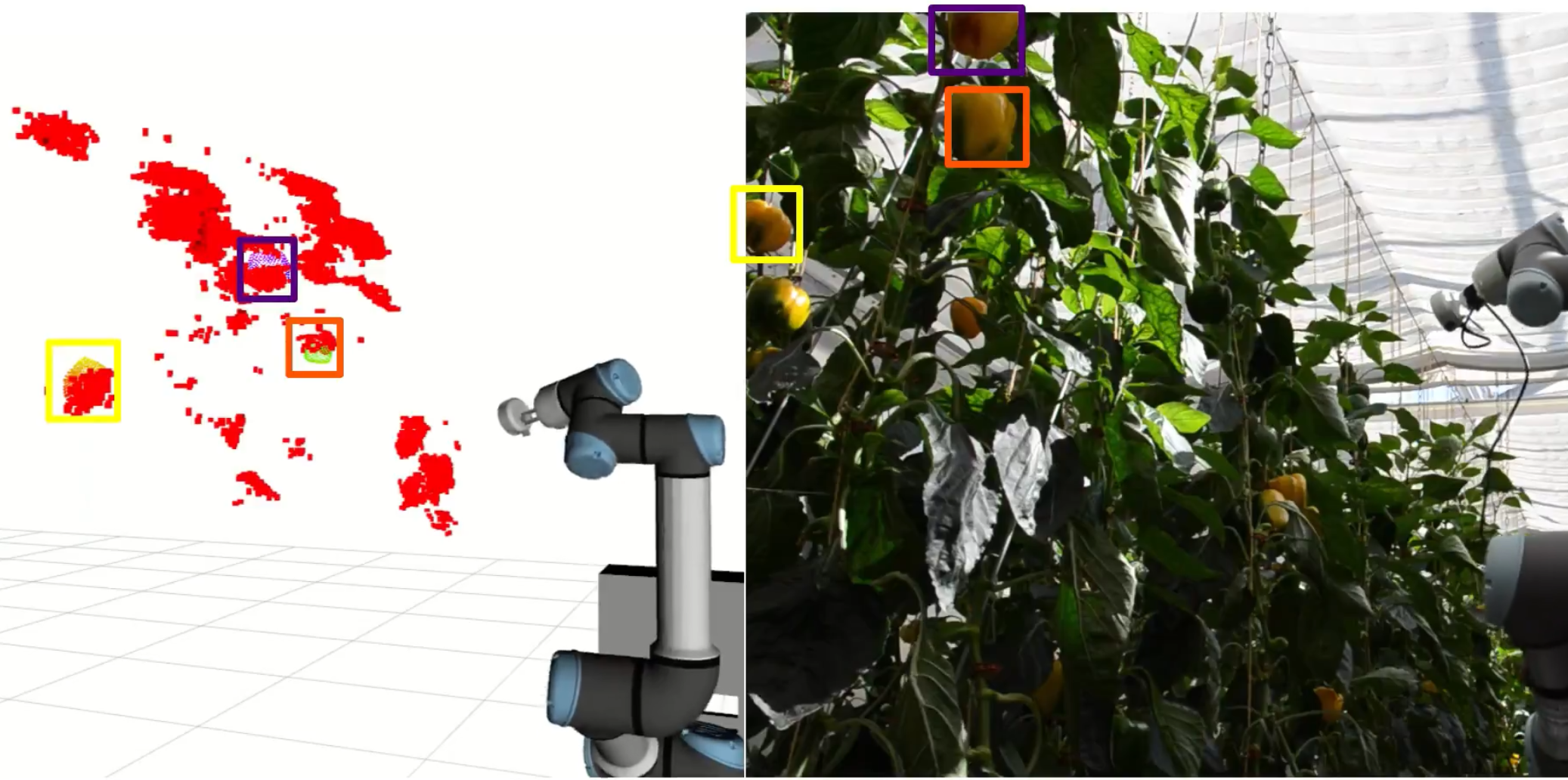} 	
	\captionsetup{width=0.99\columnwidth, justification=justified}
	\caption{NBV-SC planning viewpoints based on shape completion in a commercial glasshouse.} 
	\label{fig:real_world}
\end{figure} 
For our real-world experiments, we applied the approach to map sweet peppers in a commercial glasshouse.
We performed the experiments with a UR5e arm equipped with a Realsense L515 sensor, mounted on the trolley presented in~\cite{mccool21icra}.
An example of the performed shape completion in this environment is shown in \figref{fig:real_world}.
We provide a demonstration of the viewpoint selection in the accompanying video.

\section{Summary}
\label{sec:concl}
\textcolor{black}{Occlusion of fruits by leaves severely affects their reconstruction and volume estimation.}
In this paper, we presented a novel approach to active perception in agricultural robotics using next best view planning with shape completion feedback.
To reduce the need for computationally expensive ray casting, we adapted superellipsoid fitting based shape completion and estimate the missing surfaces to synthesize \textcolor{black}{occlusion-aware} fruit targeted viewpoints. 
We addionally formulated a new concept of viewpoint dissimilarity and demonstrated its effects on reducing the planning time.
The experimental results show that our NBV-SC planner leads to more complete coverage and thereby better reconstruction of individual fruits as well as to more accurate size estimation, while significantly reducing the planning time compared to a state-of-the-art viewpoint planner.

\bibliographystyle{IEEEtran}

\balance 
\bibliography{refs}
\balance 
\end{document}